\documentclass[preprint,review,10pt]{elsarticle}

\usepackage{csvsimple}

\usepackage{graphicx}
\usepackage{amssymb}

\usepackage{lineno}
\usepackage[colorinlistoftodos]{todonotes}

\journal{Journal Name}

\begin{document}

\begin{frontmatter}


\title{Automatic detection of lesion load change in Multiple Sclerosis  using convolutional neural networks with 
segmentation confidence}



\author[scan]{Richard McKinley}
\ead{richard.mckinley@insel.ch}
\author[scan]{Lorenz Grunder} 
\author[scan]{Rik Wepfer}
\author[scan]{Fabian Aschwanden}
\author[balgrist]{Tim Fischer}
\author[neurology]{Christoph Friedli}
\author[scan]{Raphaela Muri}
\author[scan]{Christian Rummel}
\author[tiefenau]{Rajeev Verma} 
\author[zrh]{Christian Weisstanner} 
\author[idsc]{Mauricio Reyes}
\author[neurology]{Anke Salmen}
\author[neurology]{Andrew Chan} 
\author[scan]{Roland Wiest}
\author[scan]{Franca Wagner}

\address[scan]{Support Center for Advanced Neuroimaging, University Institute for Diagnostic and Interventional Neuroradiology, Inselspital, Bern University Hospital, University of Bern, Switzerland}
\address[tiefenau]{Department of Neuroradiology, Spital Tiefenau, Switzerland}
\address[zrh]{Medizinisch Radiologischen Institut, Zurich, Switzerland}
\address[neurology]{Univeristy Clinic for Neurology,  Inselspital, Bern University Hospital, University of Bern, Switzerland}
\address[idsc]{Insel Data Science Centre, Inselspital, Bern University Hospital, University of Bern, Switzerland}
\address[balgrist]{Universit\"atsklinik Balgrist, Zurich, Switzerland}

\begin{abstract}
The detection of new or enlarged white-matter lesions in multiple sclerosis is a vital task  in the monitoring of patients undergoing disease-modifying treatment for multiple sclerosis.  However, the definition of 'new or enlarged' is not fixed, and it is known that lesion-counting is highly subjective, with high degree of inter- and intra-rater variability.  Automated methods for lesion quantification, if accurate enough, hold the potential to make the detection of new and enlarged lesions consistent and repeatable.  However, the majority of lesion segmentation algorithms are not evaluated for their ability to separate radiologically progressive from radiologically stable patients, despite this being a pressing clinical use-case.  In this paper we demonstrate that change in volumetric measurements of lesion load alone is not a good method for performing this separation, even for highly performing segmentation methods.  Instead, we propose a method for identifying lesion changes of high certainty, and establish on an internal dataset of longitudinal multiple sclerosis cases that this method is able to separate progressive from stable timepoints with a very high level of discrimination (AUC = 0.999), while changes in lesion volume are much less able to perform this separation (AUC = 0.71). Validation of the method on a second external dataset confirms that the method is able to generalize beyond the setting in which it was trained, achieving an accuracy of 83 \% in separating stable and progressive timepoints.

Both lesion volume and lesion count have previously been shown to be, together with clinical covariates, strong predictors of disease course across a population.  However, in this paper we demonstrate that for individual patients, changes in these measures are not an adequate means of establishing no evidence of disease activity. Meanwhile, directly detecting tissue which changes, with high confidence, from non-lesion to lesion  is a feasible methodology for identifying radiologically active patients.

\end{abstract}

\begin{keyword}
Deep Learning \sep Multiple Sclerosis \sep MRI \sep Longitudinal Imaging


\end{keyword}

\end{frontmatter}


\section{Introduction}

\label{S:1}

Magnetic resonance imaging is the most important imaging method for diagnosis and monitoring of multiple sclerosis. The 2017 revised Mcdonald diagnostic criteria for the diagnosis of multiple sclerosis require the dissemination of lesions in both space and time. Lesion load change is also crucial for the assessment of disease activity, since patients who are assigned with disease modifying therapies and no evidence of disease activity (NEDA) harbor a better prognosis \cite{Arnold2014, Havrdova2009,Havrdova2014,Nixon2014}.

Lesion load analysis and manual lesion segmentation by human raters is time-consuming and limited by inter- and intra-rater variability~\cite{Altay2013}. As a consequence manual lesion volumetry and lesion counting has limited sensitivity for new lesion detection. Delineation of new and enlarged lesions can be improved by working on subtraction MRI, but this still requires substantial human user interaction, as well as manual intensity normalization; furthermore, registration errors, flow artifacts and lesion signal intensity differences can result in the detection of false-positive “lesions” on subtraction images~\cite{Moraal2009}.

Several groups have proposed automated methods for multiple sclerosis lesion segmentation, mostly validated in a cross-sectional fashion.~\cite{mckinley2016,Valverde2017,Valverde2018, Fartaria2018} Even where longitudinal data was used to assess the performance of classifiers, consistency of segmentations over time, or the ability to detect new lesions were not investigated~\cite{Carass2017}.  Since MR contrast will differ between timepoints, even on the same scanner, and since the borders of MS lesions are often not well defined, automated methods will typically show small differences in the boundaries of lesions at different timepoints, even if no lesion growth has taken place.  Since even the best automated methods also make false positive and false negative lesion identifications, lesion counts may also not be reliable in a longitudinal setting.  Several researchers have proposed methods to harmonize segmentations across two or more timepoints.   Jain et al propose a joint expectation-maximization (EM) framework for two time point white matter (WM) lesion segmentation, and the Lesion Segmentation Toolkit, a tool integrated in SPM, has a longitudinal pipeline which adapts existing segmentations across multiple timepoints  \cite{Jain2016, Schmidt2012}.

In a companion paper we have introduced a novel method, DeepSCAN, based on convolutional neural networks (CNNs), for multiple sclerosis lesion segmentation, which we demonstrated to outperform previous methods.\cite{McKinley2019a}  In this paper, we demonstrate that even for such high-performing methods, neither lesion count nor lesion volume are adequate biomarkers for disease progression in multiple sclerosis.  Simultaneous lesion growth and lesion resolution may occur at a single timepoint, which will not be apparent from simply observing volume changes.  Further, variations in image contrast between acquisitions can lead to substantial volumetric changes in automated lesion delineation, even when using `state-of-the-art' classification methods.  Lesion counts are also only approximate measures of activity, since lesions may be missed or undersegmented, false positives may give the impression of lesion growth where none exists, and lesions may become confluent, leading to an increase in lesion tissue but a decrease in lesion count.

    As a potential solution to this issue, we instead propose to identify new and missing lesion tissue by using the \emph{confidence} of an automated classifier in its own segmentation. Measures of segmentation uncertainty have previously been proposed as a method of rejecting false positive MS lesion identifications.~\cite{Nair2018} To our knowledge, our method is the first to leverage segmentation confidence in the detection of longitudinal change. Our recently introduced MS lesion classifier, DeepSCAN, produces for each tissue map a 'label-flip probability', which is a measure of uncertainty derived from the training data.  We use the segmentation of the classifier and the label-flip map distinguish between patients with no new or enlarged lesions (those satisfying that component of the NEDA criteria) and those with genuinely new or enlarged lesions. We identify as new lesion tissue only those voxels that were confidently not present at timepoint t=0 but that are confidently lesion tissue at timepoint t=1. The method requires  T1, FLAIR and T2 imaging adhering to current imaging standards in MS (specifically, a 3D FLAIR and 3D T1 acquisition).~\cite{COTTON2015}.

\section{Methods}

\subsection{The DeepSCAN MS lesion classifier}
The DeepSCAN MS lesion classifier is a fully-convolutional neural network trained on fifty cases from the Bernese MS cohort databank, which provides segmentations of MS lesions, together with segmentations of the cerebellum, subcortical grey matter structures, and cortical grey and white matter.  In this study we only use the lesion segmentations produced by the classifier.

 Neural networks are trained by \emph{Stochastic Gradient Descent}, in which a loss function is minimized over many passes through the data.  For binary classification problems, such as determining whether a voxel contains lesion tissue, the typical loss function is binary crossentropy, which compares the output of the classifier to a 'ground-truth' label set: the closer the output is to the ground-truth, the lower the loss.  As we have already discussed, manual segmentations of MS lesions have large inter- and intra-rater variability, and so we must accept that this 'ground-truth' may, for lesion segmentation, contain many inconsistencies: missed or under-segmented lesions, and false identifications or over-segmented lesions.  For example, a retrospective analysis of the data used to test the DeepSCAN classifier found that am average of 18 false positive lesions and 4 missed lesions per subject.  Even where raters find the same lesions, their assessment of the border locations may differ substantially.
 
 One potential solution to this problem is to estimate the \emph{uncertainty} in the prediction coming from the (labelled) data: this is referred to by Kendall and Gal as \emph{aleatoric uncertainty}.  The DeepSCAN classifier we use in this paper was trained using a novel, hybrid loss function  which estimates this aleatoric uncertainty as a probability: the probability that the predicted label (given by the classifier) and the ground truth label (given by a human rater) differ, which we refer to as a \emph{label-flip probability}.\cite{McKinley2019c}  These probabilities range from 0 to 0.5: if this probability is close to zero, the classifier is certain of the label it predicted.  Meanwhile, a probability of 0.5  means that the output of the classifier is essentially a guess.\footnote{A label-flip probability greater than 0.5 would entail that the classifier `believes' that the label it predicted is more likely wrong than correct.  Since our goal is, as much as possible, to reproduce the output of the human rater, we do not allow flip probabilities above 0.5}  During training, voxels with high uncertainty (flip-probability close to 0.5)  contribute less to the learning process than those with low uncertainty (flip-probability close to 0). 

\subsection{Study design and overview}
In this study we utilise data from two sources.  The first are MRI datasets of patients with remitting-relapsing multiple sclerosis that were identified from the MS cohort databank of the MS cohort of the University of Bern.  Use of data for this study was approved by the local ethics committee (Cantonal Ethics Commission Bern, Switzerland 'MS segmentation disease monitoring', approval number 2016-02035) and all patients gave general consent for data storage and analysis of their MRI datasets. 

Additional data  was provided by an MR Center in Zurich (Radiology Center Bethanien), which we subsequently refer to as the Zurich dataset.  All patients gave written informed consent to participate in this study.

\subsection{Patient cohorts and MR imaging}
Patients were included if they had at least three consecutive MRI datasets and had not been used in training of the DeepSCAN classifier.~\cite{McKinley2019a}  All patients fulfilled the revised McDonald criteria of 2010 for relapsing-remitting multiple sclerosis.\cite{Polman2011} All MR images from our local MS cohort databank were acquired using a standardized acquisition protocol on a 3T MRI (Siemens Verio, Siemens, Erlangen, Germany) for patients with demyelinating CNS disorders. The protocol settings included i) T1 weighted MPR pre- and post gadobutrol i.v. (TR 2530 ms, TE 2.96 ms, averages 1, FoV read 250 mm, FoV phase 87.5 \% voxel size 1.0 x 1.0 x 1.0 mm, flip angle 7°, acquisition time 4:30 min. slices per slab 160, slice thickness of 1.0 mm)  ii) T2- weighted imaging (TR 6580 ms, TE 85 ms, averages 2, FoV read 220 mm, FoV phase 87.5 \%, voxel size 0.7 x 0.4 x 3.0 mm, flip angle 150°, acquisition time 6:03 min, 42 parallel images were acquired with a slice thickness of 3.0 mm, iii) 3D FLAIR imaging (TR 5000 ms, TE 395 ms, averages 1, FoV read 250 mm, FoV phase 100 \%, voxel size 1.0 x 1.0 x 1.0  mm, acquisition time 6:27 min. A total of 176 parallel images were acquired with a slice thickness of 1.0 mm). All patients received Gadobutrol (Gadovist™) 0.1 ml kg–1 bodyweight immediately after the acquisition of the unenhanced T1w sequence.  

MR images from the Zurich dataset were acquired using a standardized acquisition protocol on a 3T MRI (Siemens Skyra, Siemens, Erlangen, Germany).
i) T1 weighted MPRAGE precontrast (TR 2300 ms, TE 2.9 ms, TI 900 ms, averages 1, FoV read 250 mm, FoV phase 93.75 \% voxel size 1.0 x 1.0 x 1.0 mm, flip angle 9°, acquisition time 05:12 min. 176 slices per slab, slice thickness of 1.0 mm) 
ii) T2- weighted imaging (TR 4790 ms, TE 100 ms, averages 1, FoV read 220 mm, FoV phase 100 \%, voxel size 0.7 x 0.4 x 3.0 mm, flip angle 150°, acquisition time 02:16 min, 45 parallel images were acquired with a slice thickness of 3.0 mm,
iii) 3D FLAIR imaging (TR 5000 ms, TE 398 ms, TI 1800 ms, averages 1, FoV read 250 mm, FoV phase 100 \%, voxel size 1.0 x 1.0 x 1.0 mm, flip angle 120°, acquisition time 04:17 min. A total of 176 slices per slab were acquired with a slice thickness of 1.0 mm).

\subsection{Dichotomization of imaging data: progressive vs stable}
 The cohort of patients was split into those with progressive disease (PD, if any new FLAIR- or contrast-enhancing lesions was detected) or stable disease (SD, if the number of lesions remained stable or reduced over time), based on visual analysis by one of the authors (LG for cases from Bern, CW for cases from Zurich).  For the progressive cases we further recorded, for each timepoint after the baseline, if disease progression (i.e. new lesions) had been observed.  

\subsection{Automated Segmentation by DeepSCAN convolutional neural network}
For each patient and timepoint we used the DeepSCAN classifier to generate lesion masks and label-flip maps for MS lesions lesions.  To aid in comparison between timepoints, these maps were resampled to $1mm^3$ isotropic resolution.  The classifier also returns a $1mm^3$ isotropic skull-stripped FLAIR image in the same space as the lesion and label-flip maps.

\subsection{Coregistration}
In order to compare cases across timepoints, it was necessary to register all imaging for each patient to a common space. To avoid biases inherent in registering to a particular timepoint, we applied a robust registration technique in which all timepoints are registered to a common patient-specific template (the Robust Template method from Freesurfer) to the skull-stripped FLAIR images produced by our CNN tool.~\cite{Reuter2012}  After construction of the template, lesion masks and lesion confidence maps were rigidly registered to the template space using the transforms output by the robust template method.

\subsection{Lesion change detection by classification uncertainty}

We describe here the decision procedure for labelling a voxel as 'new lesion', given lesion mask and label-flip maps at timepoints A and B in a common, coregistered space, and a threshold $q$ determining acceptable confidence.  A voxel is labelled as 'confident lesion' at timepoint A if it is in the lesion mask, and if the label-flip probability is less than $q$.  A voxel is labelled 'confident non-lesion' if it is not in the lesion mask, and if the label-flip probability is less than $q$.  A voxel is labelled as 'new lesion' at timepoint B, if it is labelled as 'confident non-lesion' at timepoint A, and 'confident lesion' at timepoint B.  It is labelled 'missing lesion' at timepoint B, if it is labelled as 'confident lesion' at timepoint A, and 'confident non-lesion' at timepoint B. Finally, connected components of the 'new lesion' and 'missing lesion' maps were calculated, and all connected components containing fewer than 12 voxels were deleted. 

For the purposes of our initial investigation, we set the value of $q$ to be 0.05: i.e., we determine a voxel to be classified with confidence if the model predicts a $5\%$ or lower chance of the predicted label disagreeing with the manual rater.

\subsection{Lesion change detection by threshold margin}

A more simplistic methodology for labelling lesions as confidently or uncertainly classified is to set a margin around the ordinary decision threshold, 0.5, and to label all voxels outside of this margin as 'confident'. This method has the advantage that it may be applied to classifiers which do not output a label-flip probability: however, in general the output of modern neural networks is not well calibrated: the scores output by deep networks do not correspond to observed probabilities and are typically overconfident \cite{Guo2017}.

Concretely, we set a margin $0<m<0.5$, and classify every voxel with $p< 0.5 - m$ as confident nonlesion, while every voxel with $p> m+0.5$ is classified as confident lesion.  The measure of new lesion tissue is then as above: a voxel is new lesion if it is labelled as 'confident lesion' at timepoint A, and 'confident non-lesion' at timepoint B.  As above, connected components below 12 voxels were deleted.

For the purposes of our initial investigation, we set the value of $m$ to be 0.45: i.e., we determine a voxel to be classified as confident lesion if the model predicts a score of $.95$ or greater and to be classified as confident non-lesion if the model predicts a score of $0.05$ or less.

\subsection{Evaluation}
We compare our proposed methods to four other methods: absolute change in lesion volume, relative change in lesion volume, change in lesion count, and total new lesion volume (equivalent to our method with $q=0.5$). To test the power of these measures to separate progressive and stable timepoints, we plotted the receiver-operating characteristic (ROC) curves for each of the above methods.  We then tested the performance of our metrics at an operating threshold corresponding to 'no lesion change' (i.e. lesion count $>0$, lesion volume change $>0$, and new lesion volume $>0$). 

Finally, we assessed the sensitivity of our method to the uncertainty threshold $q$, by comparing the ROC curves of the method at different values of $q$.

\section{Results}

Twenty-six patients from our MS databank satisfied the inclusion criteria, of which 16 were judged from radiological reports to have no lesion changes in any of the timepoints, and so were labelled as having stable disease (SD). The remaining 10 cases were judged to have progressive disease (PD).  The mean number of timepoints per patient was 4.4 for the progressive patients, and 4.9 for the stable patients. Among the ten progressive patients, there were a total of 13 timepoints where the radiological reports indicated progression, meaning that approximately 30\% of the timepoints in those patients showed lesion progression.

The Zurich dataset consisted of four consectutive time-points (thirty-two datasets, twenty-four after baseline imaging).  Of the twenty-four follow-up time-points, five were judged by the rater (CW) to have new or enlarged lesions.

We tested four different metrics for detecting lesion progression, based on the DeepSCAN lesion segmentation method: absolute change in lesion volume, relative change in lesion volume, change in lesion count, and the proposed method based on uncertainty.  

\subsection{ROC-AUC analysis}

\begin{figure}
    \centering
    \includegraphics[width=12cm,
    height=12cm,
    keepaspectratio,]{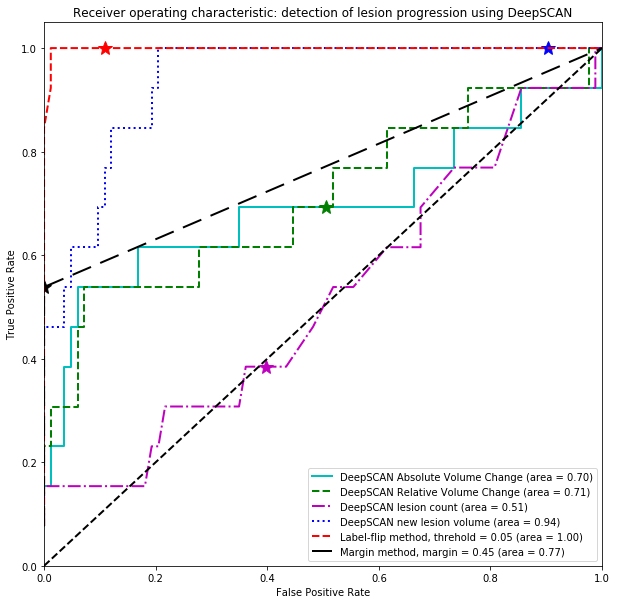}
   \caption{Receiver operating curves for the detection of lesion progression using DeepSCAN, on our internal validation set, via absolute lesion volume change (AUC=0.70), relative volume change (AUC = 0.71), lesion count change (AUC = 0.51), the proposed method using a score margin of .45 (AUC=0.77) and the proposed method using an uncertainty threshold of 0.05 (AUC $\approx 1$).  The star on each curve represents a cutoff where the patient is labelled as stable if the considered metric is less than or equal to zero.}. 
    \label{fig:rocs}
\end{figure}

For each method, we computed the area under the receiver-operating characteristic for our local cases: see Figure~\ref{fig:rocs}.  Lesion counting performed worst, with a ROC-AUC of 0.51, while absolute and relative volume change performed comparably, with ROC-AUCs of 0.70 and 0.71 respectively.  The proposed method using score margins had an AUC of 0.77.  Meanwhile, the proposed method using network-derived uncertainty had a ROC-AUC of 0.999.

\subsection{Performance at meaningful thresholds}

While ROC-AUC analysis gauges the ability of a metric to separate positive and negative examples across all operating thresholds, clinical applicability required that a particular threshold is chosen.  For the purpose of lesion progression, the obvious threshold for growth is zero: i.e. growth has occurred if there is a positive change in lesion volume/count. Results of this analysis are shown in Table~\ref{tab:my_label}.

For lesion counting, this metric leads to a total of 33 timepoints being identified as progressive, when in fact they were stable according to radiological reports.  For lesion volumetry, 42 timepoints were falsely identified as being stable.  For the proposed method, nine stable timepoints were labelled as progressive. Meanwhile, proposed method based on uncertainty successfully identified all progressive timepoints.  By comparison, the lesion volume metric failed to find four of the progressive timepoints, and lesion counting failed to find eight progressive timepoints.  The proposed method based on a margin around the decision boundary made no false positive identifications, but failed to find six of the progressive timepoints.

\begin{table}[]
    \centering
    \begin{filecontents*}{results_updated.csv}
,TN,FP,FN,TP,Accuracy,Sensitivity,PPV,FPR
Confidence method$>0$,74,9,0,13,0.91,1.00,0.59,0.11
Margin method $>0$,83,0,6,7,0.94,0.54,1.00,0.00
New lesion volume $>0$,8,75,0,13,0.22,1.00,0.15,0.90
Volume change$>0$,41,42,4,9,0.52,0.69,0.18,0.51
Lesion count change $>0$,50,33,8,5,0.57,0.38,0.13,0.40
\end{filecontents*}
\csvautotabular{results_updated.csv}
    \caption{Ability to distinguish progressive vs stable MS at  thresholds corresponding to no lesion change, on internal test set, showing the number of true negatives (TN), false positives (FP), false negatives (FN) and true positives (TP), together with accuracy, precision and recall.  Metrics are shown for the label-flip method (Confidence method) and the margin-based method (Margin method), together with new lesion volume, lesion volume change and lesion count change.}
    \label{tab:my_label}
\end{table}

\subsection{Sensitivity to uncertainty threshold, score-margin and small-growth threshold}

The best-performing method according to area under the ROC curve, according to our initial analysis,  was achieved using our uncertainty-based method with an  uncertainty threshold of $0.05$: i.e. voxels which had a flip-probability greater than 0.05 at either timepoint are not used to calculate lesion change.
At a fixed operating threshold, meanwhile, our two proposed methods performed similarly in terms of accuracy, but the method derived from label-flip confidence had perfect sensitivity and lower PPV, while the method derived from a margin around the threshold had perfect PPV and lower sensitivity.

Both of these methods rely on a parameter which can be varied, with an effect on the performance.  In this section we investigate the effect of changing those parameters.

\subsubsection{Effect of changing uncertainty threshold}
For uncertainty threshold values lower than the one we initially selected (0.0005, 0.001 and 0.01), the AUC was slightly reduced, at 0.92.  At larger uncertainty thresholds than initially selected, the AUC was also slightly lower: a threshold of 0.1 gave an AUC of 0.99, and a threshold of 0.2 gave an AUC of 0.96.

\subsubsection{Effect of changing classification margin}
The effect of changing the classification margin was much more drastic.  By setting a narrower classification margin (0.15), we were able to achieve an AUC close to the performance of the uncertainty-based method (AUC = 0.998).  A slightly larger margin of 0.2 gave worse performance (AUC = 0.96), while a slightly narrower margin of 0.1 led to a smaller decrease in performance (AUC = 0.996).

\subsubsection{Effect of changing threshold for growth}
In the method as described, areas of growth below 12 voxels do not count towards lesion growth.  The method is reasonably robust to changes in this lesion-growth threshold.  A larger threshold of 24 voxels led to an AUC of 0.96, while a smaller threshold of 6 voxels led to an AUC of 0.997.  Not applying a threshold yielded an AUC of 0.98.

\subsection{Performance on external data}
The uncertainty-based method, as described above, was applied to data from eight patients, each having four consecutive timepoints (thirty-two datsets, twenty-four after baseline). Of the twenty-four timepoints after baseline, five were labelled by the rater (CW) as having new or enlarged lesions (PD).

The proposed method successfully identified three of the five progressive timepoints and labelled an additional three incorrectly as being progressive.  The correctly labelled timepoints had new lesion loads of 3119 mm3 and 41mm3 and  39 mm3, while 29mm3, 14mm3  and 5mm3 of tissue were mislabelled as new lesion load in the three mislablelled cases.  

\begin{figure}
    \centering
    \includegraphics[width=10cm,
    keepaspectratio,]{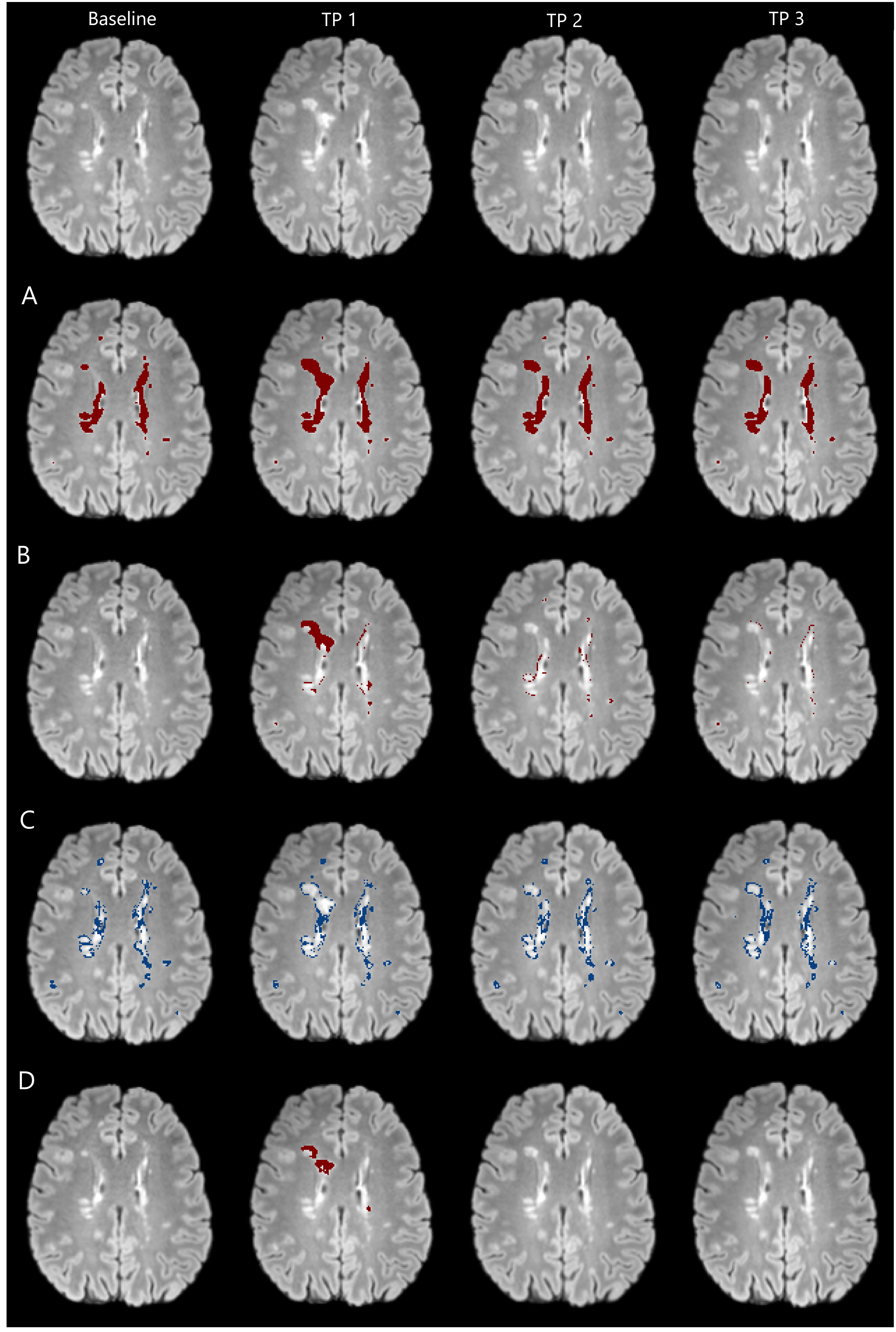}
    
    \caption{A case from the Zurich dataset.  Top Row: FLAIR imaging at baseline and three subsequent timepoints. A: FLAIR images with lesion masks as provided by the DeepSCAN classifier.
    B: FLAIR images with masks indicating naive lesion change (lesion is absent at previous timepoint but present at current timepoint). Timepoints 3 and 4 show new lesion tissue due to differences in imaging, rather than genuine lesion growth.
    C: Regions where DeepSCAN flip probability $> 0.05$ highlighted in blue.
    D: Confident new lesion tissue maps as provided by the method, showing correctly detected new lesion tissue at timepoint 2, and no change at timepoints 3 and 4}
    \label{fig:hits}
\end{figure}

\begin{figure}
    \centering
    \includegraphics[width=9cm,
    keepaspectratio,]{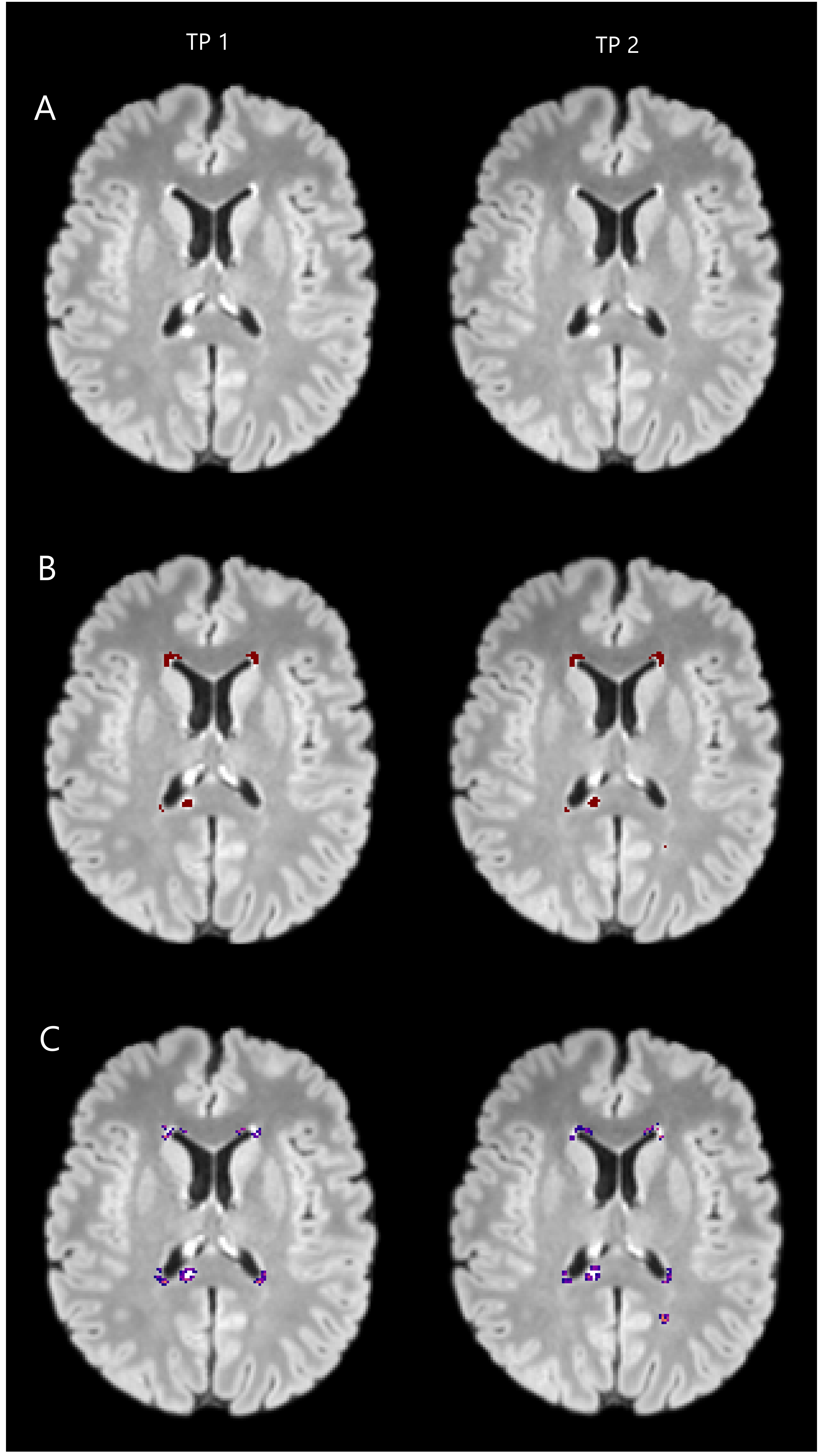}
    
\caption{Two timepoints from the external dataset, showing a missed new lesion. (A) coregistered FLAIR, (B) lesion segmentations, (C) Label-flip maps. New esion is correctly detected by DeepSCAN at TP2 , but not labelled as confident new lesion.  Small, faint lesions are more likely to be labelled as uncertain than large, clear lesions.}
    \label{fig:miss1}
\end{figure}

The two cases mislabelled as stable each had a single, small new lesion.  In the first case this was a small faint lesion in deep white matter, and in the second it was a small periventricular lesion.  In both cases these lesions were correctly segmented by DeepSCAN, but not at a sufficient level of confidence to deem them confident new lesion tissue.  Representative slices from these two cases are shown in Figures \ref{fig:miss1} and \ref{miss2}.  A representative slice from a further case from the external dataset, showing two correctly identified instances of lesion growth, is shown in Figure \ref{fig:hits}.

\begin{figure}
    \centering
    \includegraphics[width=9cm,
    keepaspectratio,]{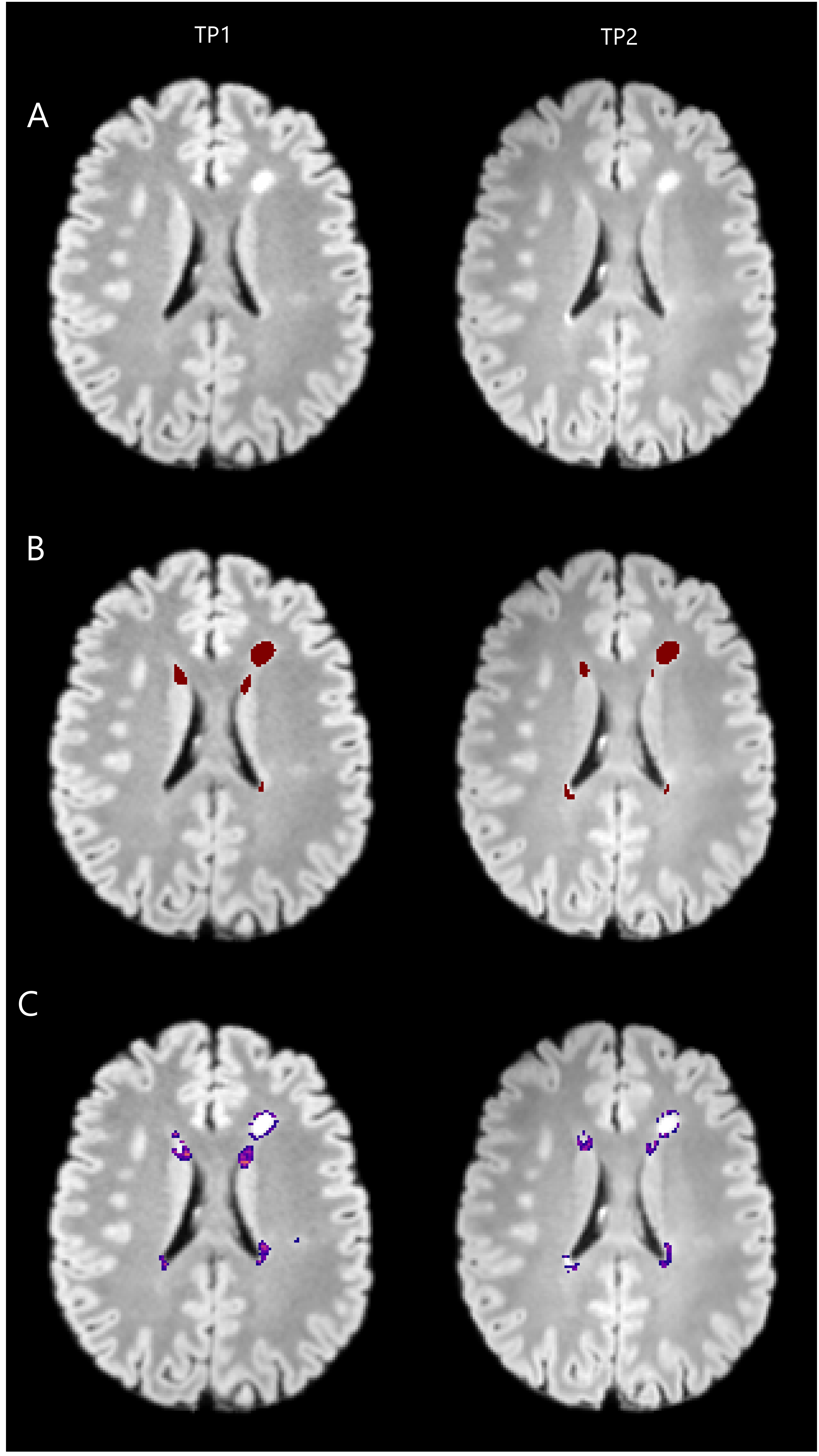}

\caption{Two timepoints from the external dataset, showing a missed new periventricular lesion. (A) coregistered FLAIR, (B) lesion segmentations, (C) Label-flip maps. Lesion is detected by DeepSCAN at TP2, but location of new lesion is uncertain at TP1.  Owing to the similar appearance of periventricular lesions and subependymal gliosis, label confidence is typically low in this region.}

    \label{fig:miss2}
\end{figure}

\section{Discussion}
MRI is the method of choice to determine lesion load evolution in patients with multiple sclerosis. The accurate detection of new or enlarged white-matter lesions in multiple sclerosis patients is a pivotal task  of the disease monitoring process in patients who receive disease-modifying treatment.  However, the definition of 'new or enlarged' remains ill-defined, and lesion counting remains subjective with a considerable degree of inter- and intra-rater variability depending on the level of experience of the reporting expert.  Automated methods for lesion quantification, if accurate, hold the potential to make the detection of new and grown lesions consistent and repeatable.  Until now, the majority of lesion segmentation algorithms are not well evaluated for their ability to accurately separate radiologically progressive disease course from radiologically stable patients during follow-up. Despite this being the pressing clinical use-case and information for the clinicians with impact on further treatment regime selection for the MS patients.   We demonstrate that measures of new lesion load derived from label-flip uncertainty outperform lesion counting as well as absolute and relative volume change detection in the longitudinal analysis of MS lesions. The major advantage of the proposed approach is to identify the timepoint during follow-up where lesion progression was evident with a very high precision. The method is fully automated, and therefore offers the benefit of being objective and independent from user bias, thus leading to more trustful longitudinal evaluations. 

The method developed relies on a minimum standard of MR imaging: in particular a 3D T1 and 3D FLAIR acquisition (with approximately $1mm^3$ or better voxels). The recommended protocol is in keeping with the 2016 Consortium of MS Centers Task Force recommendations and can be executed in approximately 20 minutes. The method in this paper proposes to track changes in lesion load by leveraging measures of uncertainty in the location of lesion boundaries, based on the predictions of a deep learning convolutional neural network classifier, DeepSCAN.  This method has already been shown to perform well at lesion segmentation in a cross-sectional setting: the classifier was more than twice as effective in lesion detection as both previous generations of CNN-based segmentation tools and freely-available lesion segmentation SPM toolboxes.~\cite{McKinley2019a} In this paper we sought to demonstrate the same classifier’s ability to detect lesion change:  by considering as new lesion tissue only those voxels which are classified confidently by DeepSCAN, progressive timepoints were detected with an accuracy of 0.91 and a recall of 1.0. By comparison with standard metrics, such as lesion count progression or volume changes, no progressive timepoints were falsely identified as stable, and the risk of false positive results decreased by more than a factor of three, in comparison with lesion counting, and a factor of eight compared to simply counting new lesion tissue voxels.  An alternative method, relying on a margin around the decision boundary rather than uncertainty, performed similarly to the label-flip confidence method, but only after the correct margin was found.  We therefore tend to prefer the uncertainty-based method.

Furthermore, our method (trained on fifty cases from a single institution) also performs well when applied to cases from a second centre: while the method failed to identify progression at two timepoints from this second centre, we can hope that such misidentifications can be avoided by training on larger, more diverse datasets.  We are also investigating the possibility to retrain our method in a semi-supervised fashion, allowing us to increase the amount and variety of data used to train the underlying models without the enormous human effort of manually labelling those datasets.

Semi-automated methods for MS lesion segmentation provide a simple method to assess the change in lesion load of an MS patient.  Simple FLAIR image subtraction methods or background subtractions of binarized image have been used to segment with high accuracy and low error rates. Other methods included graph cuts, i.e. graph-based segmentation techniques that employs seed points set by the user and a cost function or active contouring using prior information. These methods still require a degree of human interaction, are time consuming and require an “expert-in-the-loop”.   Currently, substantial effort is being invested in the development of fully-automated lesion annotation methods, and results indicate that advances in model architecture and training techniques, together with increasing availability of expert-labelled data, have brought us close to, or even allow us to exceed, the performance of expert human raters \cite{Commowick2018, McKinley2019a}.  However, in the study at hand, we could demonstrate that despite the effectiveness of automated lesion segmentation, changes in automated volumetric measurements of the lesion load in MS patients alone is not a sufficient method for performing separation between radiologically progressive course from radiologically stable patients. Instead, we propose a method for identifying lesion changes of high certainty.  We conclude that, while solitary lesion volume or total lesion load - together with clinical disease course / EDSS of MS patients - are strong predictors of disease course across a reference MS population, in the individual MS patient changes in these measures are not an adequate means to clear differentiate progressive disease course from no disease activity. 

We believe that the performance shown by our method will encourage the MS community to investigate its use in different clinical settings.  The benefits of automated methods lie not only in terms of the accuracy in differentiation of progressive versus stable disease course on MR imaging but also in the related reductions in time and economic costs derived from manual lesion labelling. While there is an increasing level of evidence that CNNs are comparable to human raters performance in cross-sectional studies, only longitudinal clinical follow-up studies will demonstrate the utility of these methods for identifying patients who remain stable under DMT.

\vspace{2em}

\section*{References}
\bibliographystyle{model1-num-names}
\bibliography{sample.bib}










\end{document}